%% file: CVPR2025/main.tex
\documentclass[10pt,twocolumn,letterpaper]{article}

\usepackage{cvpr}              

\usepackage{algorithm}
\usepackage{algpseudocode}
\usepackage{multirow}
\usepackage{booktabs}

\usepackage{color}
\usepackage{colortbl}
\usepackage{bbding}
\usepackage{pifont}

\def\method{{WeakMCN}}

\definecolor{cvprblue}{rgb}{0.21,0.49,0.74}
\usepackage[pagebackref,breaklinks,colorlinks,allcolors=cvprblue]{hyperref}

\def\paperID{3263} 
\def\confName{CVPR}
\def\confYear{2025}

\title{WeakMCN: Multi-task Collaborative Network for Weakly Supervised Referring Expression Comprehension and Segmentation}

\author{
Silin Cheng$^1$\textsuperscript{*}, Yang Liu$^2$\thanks{\scriptsize{Equal contribution.}}, 
Xinwei He$^3$, 
Sebastien Ourselin$^2$, 
Lei Tan$^4$\textsuperscript{$\mathcal{\dagger}$}, 
Gen Luo$^5$\thanks{\scriptsize{Corresponding author}
}
~\\[2mm]
$^1$The University of Hong Kong\
$^2$King's College London \
$^3$Huazhong Agricultural University \\
$^4$National University of Singapore\
$^5$OpenGVLab, Shanghai AI Laboratory}
\begin{document}
\maketitle
\input{CVPR2025/sections/0_abstract}
\input{CVPR2025/sections/1_introduction}
\input{CVPR2025/sections/2_relate_work}

\input{CVPR2025/sections/3_methodv3}

\input{CVPR2025/sections/4_experiments}

\input{CVPR2025/sections/5_conclusion}
\input{CVPR2025/sections/6_acknowledgement}
{
    \small
    \bibliographystyle{CVPR2025/ieeenat_fullname}
    \bibliography{CVPR2025/ref}
}
\input{CVPR2025/supple2}

\end{document}

%% file: CVPR2025/sections/0_abstract.tex
\begin{abstract}
Weakly supervised referring expression comprehension~(WREC) and segmentation~(WRES) aim to learn object grounding based on a given expression using weak supervision signals like image-text pairs. While these tasks have traditionally been modeled separately, we argue that they can benefit from joint learning in a multi-task framework. To this end, we propose WeakMCN, a novel multi-task collaborative network that effectively combines WREC and WRES with a dual-branch architecture. Specifically, the WREC branch is formulated as anchor-based contrastive learning, which also acts as a teacher to supervise the WRES branch. In WeakMCN, we propose two innovative designs to facilitate multi-task collaboration, namely Dynamic Visual Feature Enhancement~(DVFE) and Collaborative Consistency Module~(CCM). DVFE dynamically combines various pre-trained visual knowledge to meet different task requirements, while CCM promotes cross-task consistency from the perspective of optimization. Extensive experimental results on three popular REC and RES benchmarks, \emph{i.e.,} RefCOCO, RefCOCO+, and RefCOCOg, consistently demonstrate performance gains of WeakMCN over state-of-the-art single-task alternatives, \emph{e.g.,} up to 3.91\% and 13.11\%  on RefCOCO for WREC and WRES tasks, respectively. Furthermore, experiments also validate the strong generalization ability of WeakMCN in both semi-supervised REC and RES settings against existing methods, \emph{e.g.,} +8.94\% for semi-REC and +7.71\% for semi-RES on 1\% RefCOCO. The code is publicly available at \url{https://github.com/MRUIL/WeakMCN}.
\end{abstract}

%% file: CVPR2025/sections/1_introduction.tex
\section{Introduction}
Referring expression comprehension~(REC) and segmentation~(RES) aim to locate the target visual instance described by a referring expression, using a bounding box for localization and pixel-wise segmentation for detailed illustration~\cite{yu2016modeling,mao2016generation, kazemzadeh2014referitgame, plummer2015flickr30k}. These tasks are crucial in computer vision due to applications in various areas like human-robot interactions~\cite{cao2017realtime} and vision-language navigation~\cite{anderson2018vision}. Despite the significance, most existing methods~\cite{zhu2022seqtr,ho2022yoro, zhao2022word2pix,yang2022lavt, kim2022restr} rely on full supervision, requiring extensive fine-grained annotations that are costly and time-consuming, thereby limiting their practical applicability.
\input{CVPR2025/figures/multi_vs_single.tex}

To overcome the above limitations, weakly supervised REC (WREC) and RES (WRES) have attracted increasing attention~\cite {vaswani2017attention}. As shown in Fig.~\ref{fig:intro_diff}(a), popular WREC approaches~\cite{jin2023refclip,luo2025apl} often adopt an anchor-text matching framework to effectively leverage coarse annotations by contrastive learning. Different from WREC, WRES is typically formulated as a pseudo-label learning process~\cite{liu2023referring}. As shown in Fig.~\ref{fig:intro_diff}(b), WRES adopts a pseudo-labeling model to produce coarse-grained masks for weakly supervised learning. Despite these advancements, WREC and WRES have long been regarded as two separated tasks, and their multi-task learning is still under-explored.


\input{CVPR2025/figures/intro_frameowrk_diff.tex}
In this paper, we argue that these two tasks can be jointly learned in a single network, similar to the successful practices in fully supervised REC and RES~\cite{luo2020multi, chen2024efficient, li2021referring, su2023language, liu2023polyformer}. Nevertheless, their joint learning in a weakly supervised setting is non-trivial due to the multi-task conflict. Firstly, the modeling and learning of two tasks are distinct or even conflicting, \emph{e.g.,} contrastive learning \textit{vs.} pseudo-label learning, so directly combining WRES and WREC struggles to achieve collaborative multi-task learning. Secondly, the two tasks often pose different visual requirements, as indicated in the literature~\cite{luo2020multi}, which inevitably increases the difficulty of multi-task collaboration in a single network. Thirdly, the different task difficulties between WREC and WRES further exacerbate their inconsistency in optimization and prediction. For example, to achieve pixel-level visual-language alignment, existing WRES methods usually require more additional image-text pairs than WREC approaches~\cite{kim2023shatter, lee2023weakly, xu2022groupvit}.

To address these issues, we propose a novel multi-task collaborative network for joint WREC and WRES learning, namely WeakMCN. As shown in Fig.~\ref{fig:intro_diff}, WeakMCN formulates two tasks with a dual-branch structure, where a contrastive branch~\cite{jin2023refclip} and a multi-modal fusion branch~\cite{luo2020multi} are designed for WREC and WRES, respectively. To seamlessly connect two branches, we design an innovative cross-task pseudo-labeling method. As shown in Fig.~\ref{fig:intro_diff}, the WREC branch is optimized through the anchor-based contrastive objective, which also performs as the ``teacher'' to produce the pseudo-mask for supervising the WRES branch. By doing so, WeakMCN unifies the weakly supervised learning of two tasks into a single network. 

To encourage the collaboration of two task branches, we propose two innovative designs in WeakMCN, namely Dynamic Visual Feature Enhancement (DVFE) and Collaborative Consistency Module~(CCM). Specifically, DVFE introduces a visual bank that incorporates visual features with various pre-trained knowledge, \eg, spatial-aware knowledge in Segment Anything Model (SAM)~\cite{kirillov2023segment}. By dynamically combining features in a visual bank, DVFE can best meet the visual requirements of different task branches. In addition, CCM aims to facilitate the learning of WRES via the assistance of the WREC branch. As shown in Fig.~\ref{fig:overview}, the consistency loss and inconsistency suppression mechanism are adopted to maximize the common focus between WRES and WREC, thereby reducing the impact of unreliable pseudo masks in WRES.

To validate WeakMCN, we conduct extensive experiments on three benchmark datasets, \emph{i.e.,} RefCOCO, RefCOCO+ and RefCOCOg. As shown in Fig.~\ref{fig:multi_vs_single}, our approach consistently outperforms single-task alternatives, achieving average improvements of 7.18\% in WREC and 14.05\% in WRES across all datasets. Experimental results not only confirm the superior performance of WeakMCN than state-of-the-art (SOTA) methods on WREC and WRES, but also validate the effectiveness of its designs for multi-task collaboration. More importantly, experiments on semi-supervised REC and RES demonstrate the strong generalization ability of WeakMCN, which outperforms existing single-task SOTAs by large margins, \emph{e.g.,} +10.69\%  over RefTeacher~\cite{sun2023refteacher} on 1\% RefCOCOg for REC. In summary, our contributions are three folds:
\begin{itemize}
 \item We propose WeakMCN, a novel multi-task framework for weakly supervised Referring Expression Comprehension~(WREC) and Segmentation~(WRES) that significantly outperforms traditional single-task methods.

 \item We propose two innovative designs to facilitate the multi-task collaboration in WeakMCN: the Dynamic Visual Feature Enhancement (DVFE) for feature-wise collaboration and the Collaborative Consistency Module (CCM) for optimization-wise collaboration.

 \item Experimental results on three benchmark datasets confirm the SOTA performance of WeakMCN in both WeakREC and WeakRES, while its strong generalization ability is also validated in semi-supervised settings. 
\end{itemize}

%% file: CVPR2025/figures/multi_vs_single.tex
\begin{figure}
\centering
\includegraphics[width=1.0\linewidth]
{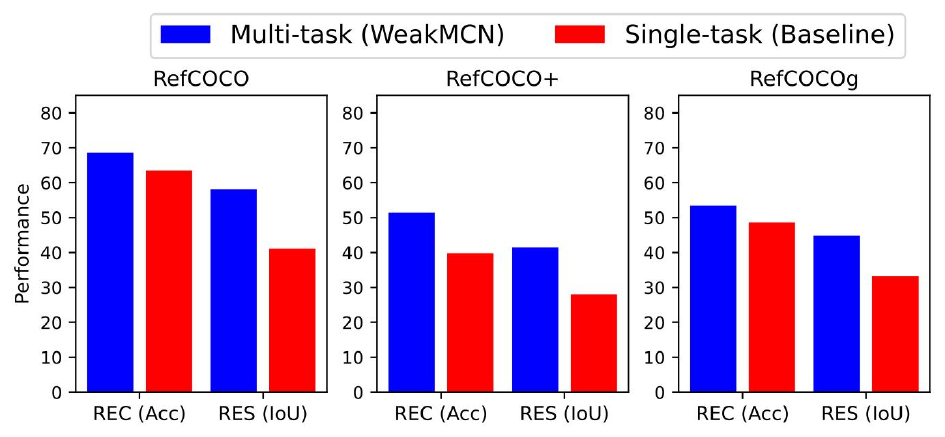}
\vspace{-1.5em}
\caption{ \textbf{Performance comparison between single-task baselines and our multi-task network (WeakMCN)}.  WeakMCN not only unifies two  tasks into a single network, but also obviously outperforms common single-task baselines. }
\label{fig:multi_vs_single}
\vspace{-2em}
\centering
\end{figure}

%% file: CVPR2025/figures/intro_frameowrk_diff.tex
\begin{figure*}
\centering
\includegraphics[width=0.95\linewidth]{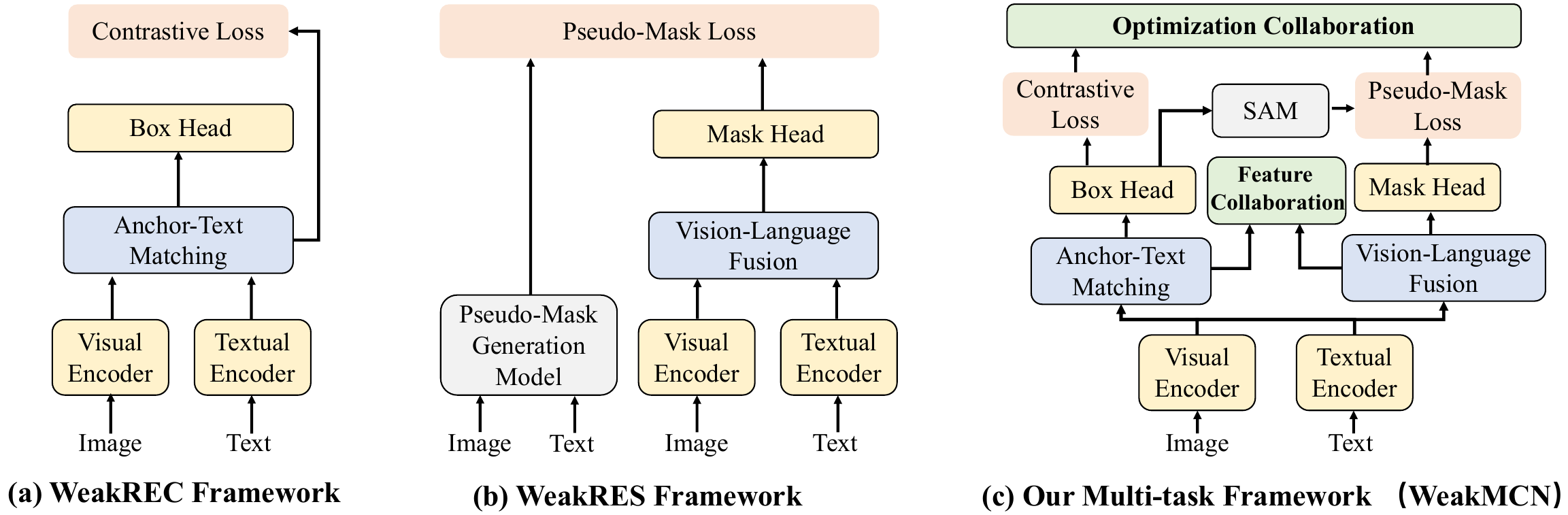}
\caption{\textbf{Comparison of previous methods and WeakMCN.}  In sub-figure (a) and (b), previous WeakREC and WeakRES often adopt the independent single-task modeling. In (c), WeakMCN is the first time to joint learn WeakREC and WeakRES in a collaborative way.}
\vspace{-1em}
\label{fig:intro_diff}
\end{figure*}

%% file: CVPR2025/sections/2_relate_work.tex
\section{Related Work}
\label{sec:related}

\subsection{Weakly Supervised REC}
 While fully supervised REC methods have achieved remarkable results~\cite{hong2019learning, zhang2018grounding, liu2019learning, liu2019improving,yang2020improving, yang2019fast, zhou2021real,liao2020real, zhou2021trar, luo2020multi,huang2021look,kamath2021mdetr, deng2021transvg, zhu2022seqtr,ho2022yoro, zhao2022word2pix, dai2024simvg},
their requirement for detailed annotations limits practical applications. This has motivated the development of weakly supervised REC (WREC) methods that rely on coarser supervision signals like image-text pairs. Early WREC methods focused on two-stage frameworks~\cite{gupta2020contrastive, liu2019adaptive, liu2019knowledge, liu2021relation, sun2021discriminative, wang2021improving, zhang2020counterfactual}, employing training objectives like sentence reconstruction~\cite{liu2019knowledge, liu2021relation, wang2021improving} and contrastive learning~\cite{gupta2020contrastive, zhang2020counterfactual}. However, these methods are computationally demanding due to the region proposal step. One-stage methods~\cite{jin2023refclip,luo2025apl,zhao2018weakly} are then focused like RefCLIP~\cite{jin2023refclip} combines anchor-text matching with contrastive learning but face challenges such as anchor ambiguity. APL~\cite{luo2025apl} improves upon this by using prompts to refine anchor representations and introducing auxiliary objectives like text reconstruction and visual alignment for better cross-modal understanding.

\subsection{Weakly Supervised RES}

Referring expression segmentation (RES) generates pixel-wise masks for target objects based on referring expressions, which require expensive pixel-level annotations~\cite{hu2016segmentation, yu2018mattnet, liu2019learning,huang2020referring, ding2021vision, yang2022lavt, kim2022restr, liu2023gres,yang2024remamber, huang2025densely}.  Instead, weakly supervised RES (WRES) methods~\cite{kim2023shatter, lee2023weakly, liu2023referring, yu2024pseudo, nag2025safari, dai2024curriculum, lyu2024gtms, yangboosting} aim to reduce the annotation burden by utilizing weaker forms of supervision, such as bounding boxes or image-text pairs. Kim~\etal~\cite{kim2023shatter} used multimodal attention to select relevant image entities for segmentation, while TRIS~\cite{liu2023referring} utilized text supervision to extract pseudo-labels for training. Lee~\etal~\cite{lee2023weakly} focused on word-level reasoning to create segmentation maps, and Dai~\etal~\cite{dai2024curriculum} used point prompting to effectively integrate SAM~\cite{kirillov2023segment}, enhancing mask quality. However, reliance on pre-trained models like SAM may still limit their application in complex scenes.

\subsection{Multitask REC and RES}

Multitask approaches~\cite{luo2020multi, chen2024efficient, li2021referring, su2023language, liu2023polyformer} aim to jointly address REC and RES by exploiting shared features between localization and segmentation tasks. MCN~\cite{luo2020multi} first introduced a multi-task collaborative network to jointly learn REC and RES. With the widespread use of Transformer-based architectures~\cite{vaswani2017attention}, follow-up works~\cite{li2021referring, su2023language} adopted a unified Transformer backbone with distinct task heads for REC and RES. Zhu~\etal~\cite{zhu2022seqtr} treated multi-task visual grounding as a sequence prediction problem, representing bounding boxes and masks as discrete coordinate tokens, while Liu~\etal~\cite{liu2023polyformer} extended this approach by using precise floating-point coordinates and generating multiple polygons for more accurate segmentation. Chen~\etal~\cite{chen2024efficient} improved upon these efforts by fusing visual and linguistic features, achieving linear scalability with respect to the expression length and reducing computational costs. These fully supervised methods benefit from the complementarity between the two tasks, achieving better overall performance. Our work extends these efforts by focusing on weakly supervised multitask learning for REC and RES~(WMRECS). We aim to reduce annotation requirements while improving accuracy by progressively integrating fine-grained attribute cues to reduce localization ambiguity and enhance segmentation precision. This approach aligns with human-like comprehension, resulting in better cross-modal alignment and overall task performance.

%% file: CVPR2025/sections/3_methodv3.tex
\section{WeakMCN}
\label{sec:method}

\input{CVPR2025/figures/Fig_overview}



In this section, we first develop a simple baseline for WMRECS. 
Based on it, we further propose two enhancing components, \emph{i.e.,} Dynamic Visual Feature Enhancement~(DVFE) and Collaborative Consistency Module~(CCM), to make the two tasks work collaboratively.

\subsection{A Simple Baseline for WMRECS}
\label{sec:baseline}


As shown in Fig.~\ref{fig:overview}, our framework consists of a multi-modal feature extractor to obtain optimal feature representations for multi-task modeling and a dual-branch structure for jointly weakly supervised learning.


\noindent{\bf Multi-modal Feature Extraction.}
As illustrated in Fig.~\ref{fig:overview}, our WREC and WRES adopt different task modeling approaches, \emph{i.e.,} contrastive learning \vs pseudo-label learning. 
In particular, we employ a shared dual-stream encoder for visual and textual feature extraction, with the visual stream outputting multi-scale features to address both tasks effectively. Specifically, for the visual stream, we utilize DarkNet from YOLOv3~\cite{redmon2018yolov3}, pre-trained on MS-COCO, to generate multi-scale feature maps $\{F_{v_i} \in \mathbb{R}^{h_i \times w_i \times d}\}_{i=1}^{3}$, which allows it to serve both tasks effectively, with spatial dimensions given by $h_i = \frac{H}{2^{i+2}}$ and $w_i = \frac{W}{2^{i+2}}$, where $H$ and $W$ are the input image dimensions. For the language stream, a bidirectional GRU encodes the referring expressions into a compact representation $f_t \in \mathbb{R}^{d_t}$, providing essential language information for two tasks.


\noindent{\bf WREC Branch.} 
For the WREC branch, we adopt an anchor-text matching mechanism to filter out the target objects inspired by RefCLIP~\cite{jin2023refclip}. Specifically, given multi-scale visual features $\{F_{v_i}\}_{i=1}^3$, we leverage only the lowest-resolution feature map $F_{v_3}$ for anchor generation, as it proves sufficient to capture referring objects in current datasets~\cite{jin2023refclip, luo2025apl}. During inference, the model predicts object locations by selecting anchors with maximum text similarity through a detection head:

\begin{equation}
    \mathbf{O_b} = \mathcal{\phi}_{\text{det}}(\arg \max_{f_v \in F_{v_3}} \langle f_v, f_t \rangle),
\label{eq:rec}
\end{equation}

\noindent where $\langle \cdot,\cdot \rangle$ computes cosine similarity between features, and $\mathcal{\phi}_{\text{det}}: \mathbb{R}^d \rightarrow \mathbb{R}^4$ represents a lightweight neural network that regresses bounding box coordinates.


\noindent{\bf WRES Branch.} Different from the anchor-based WREC branch, our WRES branch implements a multi-modal fusion strategy for pixel-wise prediction. The branch architecture is borrowed from MCN~\cite{luo2020multi}, which comprises two primary components: a multi-modal feature fusion module and a segmentation head.





During inference, the segmentation head processes fused features $F^{'}_{v_{1}}$ through a lightweight decoder comprising an ASPP module and a bilinear upsampling layer. The mask generation process follows:
\begin{equation}
    \mathbf{O_s} = \mathbb{I}[\sigma(\mathcal{U}(\text{ASPP}(F^{'}_{v_{1}}))) \geq 0.5],
\end{equation}
\noindent where ASPP$(\cdot)$ captures multi-scale context through varied dilation rates, $\mathcal{U}(\cdot)$ performs bilinear upsampling to match input resolution, $\sigma(\cdot)$ applies sigmoid activation, and $\mathbb{I}[\cdot]$ represents the thresholding indicator function. This design enables efficient end-to-end mask generation without requiring additional post-processing steps.

\noindent{\bf Joint Learning of WREC and WRES.} To enable the joint weakly supervised setting, we design task-specific losses and leverage SAM to connect the learning of two branches. For the WREC branch, we employ a contrastive learning strategy, which is formulated as: 
\begin{equation}
    L_{atc} = -\log \frac{\exp(\langle\hat{f}_{a_{i}}, f_{t_{i}}\rangle/\tau)}{\sum\limits_{j=0}^{N} \mathbb{I}(i \neq j) \exp(\langle f_{a_{j}}, f_{t_{i}}\rangle/\tau)},
\label{eq:atc}
\end{equation}
where $\hat{f}_{a_i}$ denotes the best matched anchor features in $i$-th image, \noindent where $f_{t_i} \in \mathbb{R}^d$ represents the text embedding, $N$ denotes the total number of samples in a mini-batch, $\tau$ is the temperature. 

For the WRES branch, we use $\mathbf{O_b}$ predicted by WREC as prompts for SAM to generate pseudo masks $\mathbf{\hat{M}}$. These pseudo masks then serve as supervision signals through a binary cross-entropy loss:
\begin{equation}
L_{res} = - \sum_{l=1}^{H \times W} [m_l\log (o_l) + (1-m_l)\log (1-o_l)],
\end{equation}
\noindent where $m_l$ and $o_l$ are elements of the pseudo mask $\mathbf{\hat{M}}$ and predicted mask $\mathbf{O_s}$ respectively.

\subsection{Dynamic Visual Feature Enhancement}
\label{sec:Dynamic}

WREC and WRES are two related tasks. Both of them necessitate rich features learned with broad concepts. However, they also have separate and distinct feature requirements. For instance, segmentation usually demands more fine-grained features to delineate the objects and background clearly, while detection calls for object-level features within a larger spatial context. To address these task-specific demands while leveraging their complementary nature, we propose a Dynamic Visual Feature Enhancement (DVFE) component that operates through two key mechanisms, as shown in Fig~\ref{fig:dvfe}. In essence, it enhances visual features from two following aspects: 

1) It makes use of off-the-shelf vision foundation models such as SAM~\cite{kirillov2023segment} and DINOv2~\cite{oquab2023dinov2}, which have been pre-trained on large-scale datasets with broad and diverse concepts.
This is in stark contrast with previous methods adopting DarkNet~\cite{redmon2016you} pre-trained on 80 classes of MS-COCO~\cite{lin2014microsoft}. Therefore, we can greatly excavate the potential of combining WMRECS modeling. 
In particular, given an image $I \in \mathbb{R}^{H\times W \times 3}$, we use ${N_b}$ pre-trained visual models to extract a bank of visual features $\mathcal{B} = \{V_1, \dots, V_{N_b}\}$. In our implementation, we set $N_b=3$ and DarkNet, SAM and DINOv2 are already sufficient for WMRECS modeling.

2) We perform feature selection to select appropriate features for each task separately. Specifically, for each task $t \in \{\text{`WREC'}, \text{`WRES'}\}$, we compute dynamic weights for feature combination using:

\begin{equation} 
w_{t} = \text{Softmax}(V_1\textbf{W}_t), 
\end{equation}

\noindent where $\textbf{W}_t \in \mathbb{R}^{d \times N_b}$ represents a learnable projection matrix, and $V_1$ specifically  denotes DarkNet feature ($V_1=F_{v_3}$ for WREC and $V_1=F_{v_1}$ for WRES). Then we can use the weights to adaptively ensemble the visual features to form task-specific visual feature $F_t$:

\begin{equation}
F_t = \sum_{i=1}^{N_b} w_{t,i} \cdot \mathcal{T}_{t,i}(V_i), 
\end{equation}

\noindent where $\mathcal{T}_{t,i}(\cdot)$ encompasses a linear transformation and resize operation to align features with task-specific requirements. By integrating these two complementary strategies, DVFE effectively enhances visual features for both tasks while respecting their individual requirements.

\subsection{Collaborative Consistency Module}
\label{sec:ccm}

\input{CVPR2025/figures/fig_dfve}
Multi-task learning faces the well-known multi-task conflict~\cite{luo2020multi} challenge during optimization. 
When unifying WRES and WREC, it is essential to balance the two tasks carefully, as WRES which requires pixel-wise prediction is generally more difficult than WREC.  
To address these issues, a novel Collaborative Consistency Module (CCM), which includes two innovative designs called Spatial Consistency Loss (SCL) and Inconsistency Suppression Loss (ISL), as shown in Fig~\ref{fig:ccm}.

\input{CVPR2025/figures/fig_ccm}

\input{CVPR2025/tables/RECS_sota}

\noindent{\bf Spatial Consistency Loss.}  The core idea of SCL is to facilitate the learning of WRES with the help of WREC's better grounding ability. Inspired by prior work in object detection~\cite{tian2021boxinst}, we transform the prediction of WRES and WREC into the binary distribution on x and y axes, and then compute the 1-D alignment loss. We use the bounding box predicted by the WREC branch rather than using the ground truth.
Specifically, Let $\mathbf{\hat{M}_c} \in \{0,1\}^{H \times W}$ be a binary mask derived from the predicted bounding box. We define projection operators $\mathcal{P}_x$ and $\mathcal{P}_y$ that project 2D masks onto x- and y-axes respectively:

\begin{equation}
\begin{aligned}
    \mathcal{P}_x(M)[j] &= \max_{i \in [1,H]} M[i,j], \\
    \mathcal{P}_y(M)[i] &= \max_{j \in [1,W]} M[i,j],
\end{aligned}
\label{eq:projection}
\end{equation}

\noindent where $M \in \mathbb{R}^{H \times W}$ represents the input mask, and $[i,j]$ denotes the element at the $i$-th row and $j$-th column. The Spatial Consistency Loss is then formulated as:
\small
\begin{equation}
    L_{scl} = L_{dc}(\mathcal{P}_x(\mathbf{O_s}), \mathcal{P}_x(\mathbf{\hat{M}_c})) + L_{dc}(\mathcal{P}_y(\mathbf{O_s}), \mathcal{P}_y(\mathbf{\hat{M}_c})),
    \label{eq:proj}
\end{equation}
\normalsize
\noindent where the Dice loss $L_{dc}$ between two vectors $\mathbf{p}$ and $\mathbf{q}$ is defined as:
\begin{equation}
L_{dc}(\mathbf{p}, \mathbf{q}) = 1 - \frac{2\sum_i p_i q_i}{\sum_i p_i^2 + \sum_i q_i^2 + \epsilon},
\label{eq:dice}
\end{equation}
\noindent where $\epsilon$ is a small constant to ensure numerical stability.



\noindent{\bf Inconsistency Suppression Loss.}~While SCL enforces consistency between the outputs of the WREC and WRES heads, the performance of WRES remains limited by the quality of pseudo labels. Due to limitations in SAM, the generated pseudo masks may not always align with the bounding box predictions, particularly when other objects are present within the bounding box, leading to incorrect segmentation results.

To address the above issue, we further introduce an Inconsistency Suppression Loss (ISL) to mitigate the impact of noisy pseudo masks on WRES training.
Specifically, we compute Intersection over Union (IoU) between the predicted mask $\mathbf{O_s}$ and a bounding box mask $\mathbf{\hat{M}_c}$ and leverage it as a measure of consistency between outputs of both branches. Then samples with an IoU below a threshold $\alpha$ are excluded from the segmentation loss calculation:
\begin{equation} L_{inc} = \mathbb{I}[\text{IoU}(\hat{\mathbf{O_s}} , \mathbf{\hat{M}_c}) \geq \alpha] \cdot L_{res}, \label{eq
} \end{equation}

\noindent where $\alpha$ is a quality threshold, $\mathbb{I}[\cdot]$ is the indicator function.
This formulation ensures that segmentation loss is only applied when there is sufficient alignment between the predicted mask and the bounding box, thereby reducing the negative impact of inconsistent pseudo labels.

\subsection{Overall Loss}




The overall training objective combines losses from the WREC branch, WRES branch, and the CCM:

\begin{equation} L_{total} = \lambda_{atc}L_{atc} + \lambda_{res}L_{res} + \lambda_{inc}L_{inc} + \lambda_{scl}L_{scl}, \label{eq
} \end{equation}

\noindent where the coefficients $\lambda_{atc}$, $\lambda_{res}$, $\lambda_{inc}$, and $\lambda_{scl}$ are hyperparameters that balance the contributions of each loss term.

%% file: CVPR2025/figures/Fig_overview.tex
\begin{figure*}
\centering
\includegraphics[width=0.9\linewidth]{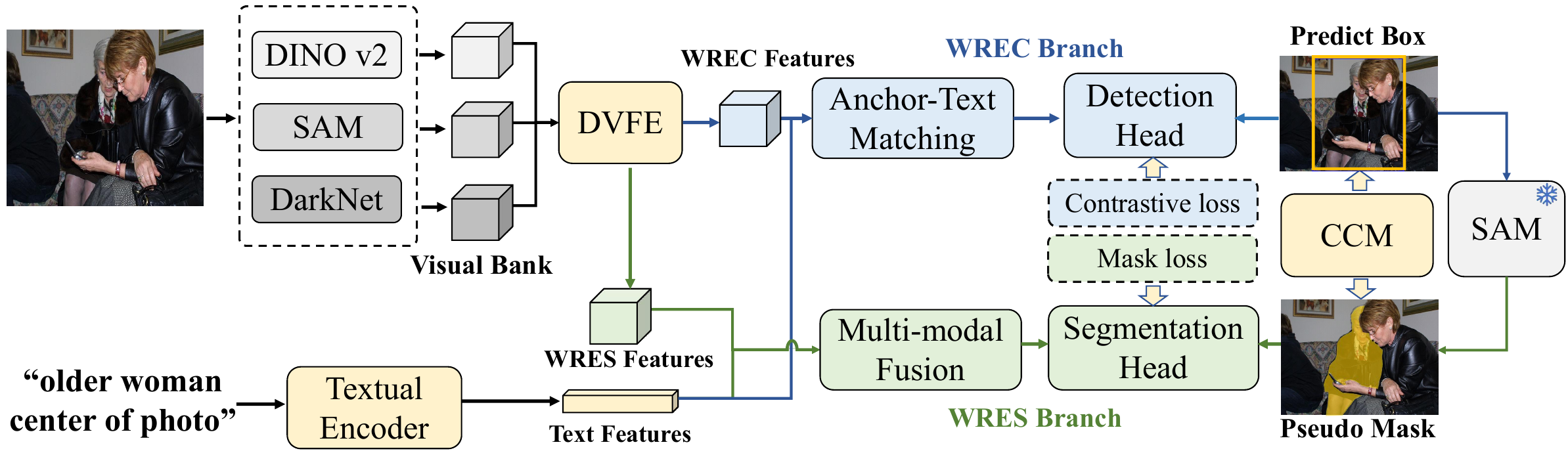}
\caption{\textbf{The overall framework of \method{}.} The referring expression is processed by a text encoder, while the image is processed by multiple foundation models and aggregated into a visual bank. The DVFE module dynamically retrieves features from this visual bank to support the WREC and WRES branches, which predict the target bounding box and segmentation mask, respectively. During training, contrastive loss and SAM-based pseudo-labeled mask loss are used, with the CCM module enhancing collaboration between both tasks.
}
\label{fig:overview}
\centering
\end{figure*}

%% file: CVPR2025/figures/fig_dfve.tex
\begin{figure}
\centering
\includegraphics[width=0.9\linewidth]{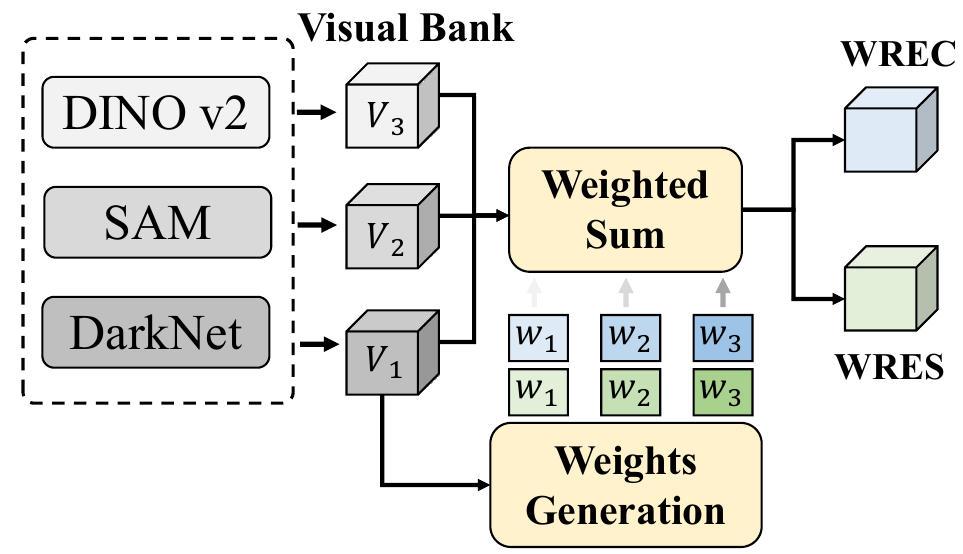}
\caption{\textbf{Overview of Dynamic Visual Feature Enhancement~(DVFE). } DVFE predicts two groups of weights to dynamically combine visual features for WREC and WRES. 
}
\label{fig:dvfe}
\centering
\end{figure}

%% file: CVPR2025/figures/fig_ccm.tex
\begin{figure}
\centering
\includegraphics[width=\linewidth]{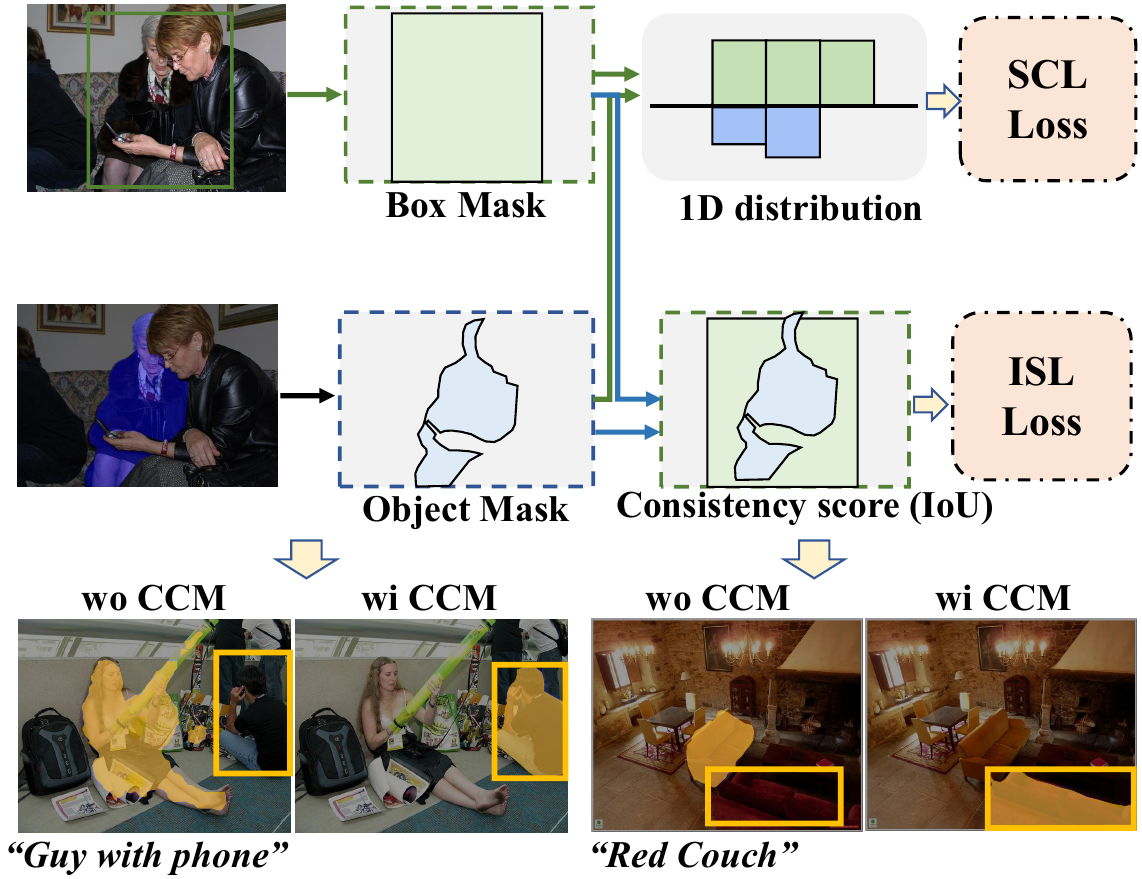}
\caption{\textbf{The Collaborative Consistency Module~(CCM) architecture.} It consists of a Spacial Consistency Loss~(SCL) $L_{scl}$ and an Inconsistency Suppression Loss~(ISL) $L_{inc}$.
}
\label{fig:ccm}
\centering
\end{figure}

%% file: CVPR2025/tables/RECS_sota.tex
\begin{table*}[h]
    \centering
    \caption{\textbf{Quantitative comparison with state-of-the-art models in REC, WREC and WRES on RefCOCO, RefCOCO+, and RefCOCOg.} F denotes fully supervision and T denotes text-only supervision. `$\dag$' indicates WeakMCN with SAM ViT-base backbone for fair comparison with~\cite{dai2024curriculum}, while `$*$' denotes WeakMCN with SAM ViT-tiny backbone.}
    \resizebox{0.95\textwidth}{!}{
    \begin{tabular}{c|l|c|c|c|c|ccc|ccc|c}
        \toprule
        \multirow{2}{*}{Task} & \multirow{2}{*}{Method} & \multirow{2}{*}{Published on} & \multirow{2}{*}{Supervision} & \multirow{2}{*}{Extra Image-text Pairs} & \multirow{2}{*}{Multi-Task} & \multicolumn{3}{c|}{RefCOCO} & \multicolumn{3}{c|}{RefCOCO+} & RefCOCOg \\
        && & & & & val & test A & test B & val & test A & test B & val-g \\
        \midrule
        \multirow{5}{*}{REC}
        &HiVG~\cite{xiao2024hivg} &ACM MM '24 & F & \ding{51} & \ding{55} & 88.14 & 91.09 & 83.71 & 80.10 & 86.77 & 70.53 & -\\ 
        &RefFormer~\cite{wangreferencing} &NurIPS '24 & F & \ding{55} & \ding{55} & 86.52 & 90.24 & 81.42 & 76.58 & 83.69 & 67.38  & - \\
        &SimVG~\cite{dai2024simvg} &NurIPS '24 & F & \ding{55} & \ding{55} & 90.61 & 92.53 & 87.68 & 85.36 & 89.61 & 79.74 & 79.34 \\
        &OneRef~\cite{xiao2024oneref} &NurIPS '24 & F & \ding{51} & \ding{51} &92.87 &94.01 &90.19 &87.98 &91.57 &83.73 & - \\
        &C$^3$VG~\cite{dai2025multi} &AAAI '25 & F & \ding{55}  & \ding{51} &92.51 &94.60 &88.71 &87.44 &90.69 &81.42 & - \\
        \midrule
        \multirow{8}{*} {WREC}&VC~\cite{niu2019variational} & CVPR '18 & T & \ding{55} & \ding{55} & - & 32.68 & 27.22 & - & 34.68 & 28.10 & 29.65 \\
        &ARN~\cite{liu2019adaptive} & ICCV '19 & T & \ding{55} & \ding{55} & 32.17 & 35.25 & 30.28 & 32.78 & 34.35 & 32.13 & 33.09 \\
        &IGN~\cite{zhang2020counterfactual} & NeurIPS '20 & T & \ding{55} & \ding{55} & 34.78 & - & - & 36.91 & 36.91 & 35.46 & 34.92 \\
        &DTWREG~\cite{sun2021discriminative} & TPAMI '21 & T & \ding{55} & \ding{55} & 38.35 & 39.51 & 37.01 & 38.19 & 39.91 & 37.09 & 42.54 \\
        &RefCLIP~\cite{jin2023refclip} & CVPR '23 & T & \ding{55} & \ding{55} & 60.36 & 58.58 & 57.13 & 40.39 & 40.45 & 38.86 & 47.87 \\
        &APL~\cite{luo2025apl} & ECCV '24 & T & \ding{55} & \ding{55} & 64.51 & 61.91 & \textbf{63.57} & 42.70 & 42.84 & 39.80 & 50.22 \\
        \rowcolor{gray!20}
         &WeakMCN$^*$ & - & T & \ding{55} & \ding{51} & \underline{68.55} & \textbf{70.78} & 62.00 & \underline{51.48} & \underline{56.92} & \underline{41.75} & \underline{53.44} \\
        \rowcolor{gray!20}
         &WeakMCN$^\dag$  & - & T & \ding{55} & \ding{51} & \textbf{69.20} & \underline{69.88} &  \underline{62.63} & \textbf{51.90} & \textbf{57.33} & \textbf{43.10} & \textbf{54.62} \\
        \bottomrule
        \multirow{3}{*}{RES} 
        & DETRIS~\cite{huang2025densely} &AAAI '25 & F & \ding{55}  & \ding{55} & 77.30 & 79.00 & 75.20 & 70.80 & 75.30 & 64.70 & 67.90 \\
        &OneRef~\cite{xiao2024oneref} &NurIPS '24 & F & \ding{51}  & \ding{51} &80.09 &82.19 &77.51 &75.17 &79.38 &70.17 & - \\
        &C$^3$VG~\cite{dai2025multi} &AAAI '25 & F & \ding{55}  & \ding{51} &81.37 &82.93 &79.12 &77.05 &79.61 &72.40 &- \\
        \midrule
        \multirow{10}{*}{WRES}&GroupViT~\cite{xu2022groupvit} & CVPR '22 & T & \ding{51} & \ding{55} & 12.97 & 14.98 & 12.02 & 13.31 & 15.08 & 12.41 & 16.84 \\
        &TSEG~\cite{strudel2022weakly} & arXiv '22 & T & \ding{51} & \ding{55} & 25.44 & - & - & 18.22 & - & - & 22.05 \\
        &ALBEF~\cite{li2021align} & NeurIPS '21 & T & \ding{51} & \ding{55} & 23.11 & 22.79 & 23.42 & 22.44 & 22.07 & 22.51 & 24.18 \\
        &TRIS~\cite{liu2023referring} & ICCV '23 & T & \ding{51} & \ding{55} & 31.17 & 32.43 & 29.56 & 30.90 & 30.42 & 30.80 & 36.00 \\
        &Chunk~\cite{lee2023weakly} & ICCV '23 & T & \ding{51} & \ding{55} & 31.06 & 32.30 & 30.11 & 31.28 & 32.11 & 30.13 & 32.88 \\
        &Shatter~\cite{kim2023shatter} & ICCV '23 & T & \ding{51} & \ding{55} & 34.76 & 34.58 & 35.01 & 28.48 & 28.60 & 27.98 & 28.87 \\
        &PPT~\cite{dai2024curriculum} & CVPR '24 & T & \ding{51} & \ding{55} & 46.76 & 45.33 & 46.28 & \textbf{45.34} & 45.84 & \textbf{44.77} & 42.97 \\
        \rowcolor{gray!20}
         &WeakMCN$^*$ & - & T & \ding{55} & \ding{51} & \underline{58.15} & \underline{59.43} & \underline{55.85} & 41.48 & \underline{46.80} & 34.94 & \underline{44.83} \\
        \rowcolor{gray!20}
         &WeakMCN$^\dag$ & - & T & \ding{55} & \ding{51} & \textbf{59.26} & \textbf{61.18} & \textbf{57.25} & \underline{44.97} & \textbf{50.83} & \underline{37.39} & \textbf{46.90} \\
         \bottomrule
    \end{tabular}
    }
    \label{tab:recs-sota}
\end{table*}

%% file: CVPR2025/sections/4_experiments.tex
\section{Experiment}
\label{sec:exp}

\subsection{Experimental Design}

\noindent\textbf{Datasets.}~We evaluate our approach on three benchmark datasets derived from MS-COCO~\cite{lin2014microsoft}: RefCOCO~\cite{yu2016modeling}, RefCOCO+~\cite{yu2016modeling}, and RefCOCOg~\cite{nagaraja2016modeling}. These datasets present diverse challenges in referring expression comprehension and segmentation. RefCOCO contains 142,210 referring expressions for 50,000 objects across 19,994 images, with separate test sets (testA and testB) focusing on person and non-person objects, respectively. RefCOCO+ comprises 141,564 expressions for 49,856 objects in 19,992 images, emphasizing appearance attributes while excluding absolute spatial references. RefCOCOg features 95,010 expressions (average length: 8.4 words) describing 49,822 objects in 25,799 images, incorporating both appearance attributes and spatial relationships. Following previous methods~\cite{jin2023refclip, luo2025apl, dai2024curriculum, kim2023shatter}, we adopt the Google split~\cite{nagaraja2016modeling} for weakly-supervised evaluation.

\noindent\textbf{Training details.} The default visual encoder is DarkNet~\cite{redmon2018yolov3}, which is borrowed from RefCLIP~\cite{jin2023refclip}. Furthermore, we also add DINOv2~\cite{oquab2023dinov2} and efficientSAM~\cite{xiong2024efficientsam} in DVFE. Input images are resized to $416\times416$ and the text embedding is initialized by GLOVE~\cite{pennington2014glove}, with maximum sequence lengths of 15 for RefCOCO/RefCOCO+ and 20 for RefCOCOg. For text encoder, we use a GRU with 1,024-dimensional hidden states. In REC branch, both anchor and text features are projected to 512-dimensional space for contrastive learning. The WRES branch adopts efficientSAM~\cite{xiong2024efficientsam} to produce the pseudo-mask. During training, optimize is set to Adam~\cite{kingma2014adam} with an initial learning rate of $1\times10^{-4}$ and batch size 64. Training proceeds for 25 epochs with cosine learning rate decay. The loss weights $\lambda_{atc}$, $\lambda_{inc}$ and $\lambda_{scl}$ are set as 1, 50 ,1 respectively. Other configurations align with RefCLIP settings.

\noindent\textbf{Metrics.}~For WREC, we use IoU@0.5 as the metric. In particular, a prediction is considered correct when the IoU between the prediction and the ground truth is larger than 0.5. For WRES, we use mIoU as the metric, which averages the IoU scores of all testing samples.

\subsection{Results of WRES and WREC}
In Tab.~\ref{tab:recs-sota}, we compare \method{} with SOTA methods across all dataset partitions, demonstrating that our method achieves quite promising results under the same level of supervision (WREC and WRES). For WREC (upper part of Tab.~\ref{tab:recs-sota}), our \method{} outperforms state-of-the-art model (APL~\cite{luo2025apl}) by +3.91\%, +9.00\%, and +4.40\% on RefCOCO, RefCOCO+, and RefCOCOg, respectively. 
In the WRES setting (lower part of Table~\ref{tab:recs-sota}
), all compared methods use extra image-text pairs, whereas our \method{} does not. Despite this, \method{} demonstrates considerable average mIoU gains over the best existing method PPT~\cite{dai2024curriculum} by +13.11\% and +8.93\% on RefCOCO and RefCOCOg, respectively. On RefCOCO+, \method{} maintains comparable performance with only a slight decrease of -2.76\%. Additionally, \method{} outperforms TRIS~\cite{liu2023referring} on RefCOCOg by +4.57\%, surpassing the previous best performance on this dataset. These results demonstrate that our collaborative design better integrates detection and segmentation, ensuring superior and consistent performance. Unlike previous methods focusing on either WREC or WRES, \method{} is the only approach that integrates both tasks effectively in a multi-task framework, enhancing consistency between localization and segmentation while improving overall quality.
\input{CVPR2025/tables/Ablation_component}
\input{CVPR2025/tables/Baseline_comparison}



\subsection{Ablation Study}
To comprehensively evaluate the effectiveness of our proposed \method{WeakMCN}, we conduct ablation studies on the \textit{val} set of RefCOCO and RefCOCO+ using SAM with ViT-Base image encoder as our default configuration.

\noindent \textbf{Different Components of the Model.}~Tab.~\ref{tab:res_componet} shows the ablation study on our proposed \method. The baseline model (second row) integrates both WREC and WRES heads, enabling multi-task capabilities. However, there is a slight decline in WREC performance compared to the original RefCLIP with an average decrease of -0.54\%, likely due to feature competition between the two tasks without an effective collaboration mechanism. Adding DVFE (fourth row) significantly improves both tasks. WREC accuracy increases by +4.84\% on RefCOCO, with an additional +0.86\% gain compared to the single-task setup, highlighting DVFE's ability to alleviate feature competition and promote effective collaboration. The WRES task also shows notable gains, with an average mIoU increasing by +9.1\%. Introducing CCM, consisting of SCL and ISL, further enhances consistency between WREC and WRES. SCL improves spatial alignment, boosting mIoU to 55.47 on RefCOCO, while ISL further refines WRES quality, increasing mIoU to 58.15. Additionally, ISL slightly benefits WREC, demonstrating the positive impact of WRES consistency on WREC.

\input{CVPR2025/tables/Rich_vision}
\input{CVPR2025/tables/semi-sota}

\input{CVPR2025/figures/Visualization_cases}
\noindent\textbf{Baseline Comparison.}
To validate WeakMCN's ability in promoting collaborative learning between WREC and WRES, we conduct comparisons against single- and multi-task baselines. As shown in Tab.~\ref{tab:baseline_comparison}, a naive multi-task architecture without collaborative mechanisms exhibits marginal performance degradation compared to single-task baselines, due to the competing optimization objectives during joint training. In contrast, by incorporating our proposed DVFE and CCM to facilitate task interaction, WeakMCN achieves substantial improvements of 8.89\% and 12.50\% in WREC and WRES tasks respectively, demonstrating the effectiveness of our collaborative design in promoting mutual enhancement between the two tasks.

\noindent \textbf{The Analysis of DVFE.}~Tab.~\ref{tab:rich_vision} shows the impact of incorporating visual bank features and adaptive selection within DVFE. The first row presents our multi-task baseline without DVFE, which employs DrakNet pre-trained on 80 object classes from MS-COCO as its sole visual encoder. By incorporating richer visual bank features, such as DINOv2 ($V_{dino}$) and SAM ($V_{sam}$), leads to notable improvements across both WRES and WREC tasks. Adding $V_{dino}$ alone significantly boosts WRES, while incorporating both DINOv2 and SAM provides further gains. Adaptive selection yields the most significant gains for both tasks. With the adaptive selection, $V_{dino}$ alone improves WREC to 67.37 (+2.71\%) and WRES to 56.14 (+3.52\%) on RefCOCO. The combination of $V_{dino}$ and $V_{sam}$ with adaptive selection achieves the best results, with WREC of 68.55 and WRES of 58.15 on RefCOCO, reflecting a clear advantage of dynamic feature adaptation over static aggregation.


\subsection{Generalizations to Semi-REC and Semi-RES}

We further evaluate WeakMCN under semi-supervised settings with only 1\% of labeled data, with SViT-tiny as the SAM image encoder. Unlike existing single-task methods, our approach enables joint learning of both tasks with limited supervision. For Semi-REC, WeakMCN surpasses RefTeacher~\cite{sun2023refteacher} by over 10\% mIoU on average across RefCOCO, RefCOCO+, and RefCOCOg, with maximum improvement of 13\% on RefCOCO+. For Semi-RES, WeakMCN outperforms SemiRes~\cite{yang2024sam} by more than 8\% mIoU. These results validate the strong generalization capability of our method in semi-supervised scenarios.

\subsection{Qualitative Results}
To gain deep insights into WeakMCN, we visualize its predictions in Fig.~\ref{fig:visualization}. Specifically, the comparative studies in Fig.~\ref{fig:cases-sota} demonstrate that our model is better in understanding complex spatial relationships and fine-grained visual attributes than its single task counterpart. This advantage can be attributed to our collaborative consistency architecture, which facilitates aligned feature learning and maintains prediction consistency between WREC and WRES tasks. Moreover, our ablation studies (Fig.~\ref{fig:cases-ablation}) further validate the crucial role of each component, where removing the CCM module leads to inconsistent predictions between detection and segmentation tasks, while excluding the DVFE module significantly impairs the model's ability to capture fine-grained visual-linguistic correlations. These findings emphasize the complementary nature of our designed modules in achieving robust performance. 

%% file: CVPR2025/tables/Ablation_component.tex
\begin{table}[t]
    \centering
    \caption{ \textbf{Ablation studies of  \method{}.} We report results on  \textit{val} set of RefCOCO and RefCOCO+.}
    \vspace{-0.5em}
    \resizebox{0.45\textwidth}{!}{
    \begin{tabular}{cccc|cc|cc}
        \toprule
         \multirow{2}{*}{WRES} &\multirow{2}{*}{DVFE}
         &\multicolumn{2}{c|}{CCM} & \multicolumn{2}{c|}{RefCOCO} & \multicolumn{2}{c}{RefCOCO+} \\
         & &SCL & ISL &REC &RES &REC &RES \\
         \midrule
           &  &  &  &63.52  &-  &39.82  &-\\
         \ding{51} &  &  &  &62.89  &45.27  &39.37 &27.91\\
           &\ding{51} &  &  & 67.36 & - &48.94 &-\\
         \ding{51} &\ding{51} &  &  &68.22  &54.08  &50.43  &37.31\\
         \ding{51} &\ding{51} &\ding{51} &  &68.33  &55.47  &51.06  &40.77\\
         \ding{51} &\ding{51} &\ding{51} &\ding{51} &68.55  &58.15  &51.48  &41.48\\
        \bottomrule
    \end{tabular}
    }
    \label{tab:res_componet}
\end{table}

%% file: CVPR2025/tables/Baseline_comparison.tex

\begin{table}[t]
    \centering
    \caption{ \textbf{Comparison of WeakMCN with single- and multi-task baselines.} ``SingleWREC'' adopts the RefCLIP as the main structure. ``SingleWRES'' uses RefCLIP and SAM to generate pseudo-masks for WRES training.}
    \vspace{-0.5em}
    \resizebox{0.45\textwidth}{!}{
    \begin{tabular}{l|cc|cc}
        \toprule
         \multirow{2}{*}{Model} & \multicolumn{2}{c|}{RefCOCO} & \multicolumn{2}{c}{RefCOCO+} \\
         &REC&RES &REC &RES \\
         \midrule
         SingleWREC &63.52 &- &39.82 &-\\
         SingleWRES &- & 46.17  &- & 28.47 \\
         \midrule
         SingleWREC+RES head (baseline) & 62.89 &45.27 &39.37 & 27.91\\
         WeakMCN &68.55 &58.15 &51.48 &41.48\\
        \bottomrule
    \end{tabular}
    }
    \vspace{-1em}
    \label{tab:baseline_comparison}
\end{table}

%% file: CVPR2025/tables/Rich_vision.tex
\begin{table}[t]
    \centering
    \caption{\textbf{Ablation studies of DVFE in WeakMCN.}}
    \vspace{-0.5em}
    \resizebox{0.45\textwidth}{!}{
    \begin{tabular}{ccc|c|cc|cc}
        \toprule
         \multicolumn{3}{c|}{$\mathcal{B}$} &\multirow{2}{*}{Adpt.}& \multicolumn{2}{c|}{RefCOCO} & \multicolumn{2}{c}{RefCOCO+} \\
         $V_{dark}$&$V_{dino}$&$V_{sam}$ &&REC&RES &REC &RES \\
         \midrule
        \ding{51} & & & &63.95 &46.88 &39.84 &28.61 \\
        \ding{51} & \ding{51} & & &64.10 &53.64 &47.09 &36.90 \\
        \ding{51} &\ding{51} & \ding{51} & &64.66 &56.46 &48.78 &37.91 \\
         \midrule
         \ding{51} &\ding{51} & & \ding{51} &67.37 &56.14 &50.32 &40.43 \\
        \ding{51} &\ding{51}&\ding{51} & \ding{51} &68.55 &58.15 &51.49 &41.47 \\
        \bottomrule
    \end{tabular}
    }
    \label{tab:rich_vision}
\end{table}

%% file: CVPR2025/tables/semi-sota.tex
\begin{table}[t]
    \centering
    \caption{ \textbf{Comparison with existing methods on Semi-REC and Semi-RES.} We use SAM with ViT-tiny backbone. ``MT'' refers to multi-task learning. For WeakMCN, we  use 1\% labeled object center point for anchor-based contrastive learning.}
    \vspace{-0.5em}
    \resizebox{0.48\textwidth}{!}{
    \begin{tabular}{l|c|ccc|ccc|c}
        \toprule
        \multirow{2}{*}{Method}  & \multirow{2}{*}{MT} & \multicolumn{3}{c|}{RefCOCO} & \multicolumn{3}{c|}{RefCOCO+} & RefCOCOg \\
         & & val & test A & test B & val & test A & test B & val-g \\
        \midrule
        \multicolumn{9}{l}{Semi-REC (1\% labeled data)}\\
        \midrule
        RefTeacher~\cite{sun2023refteacher} &\ding{55} & 59.25 & 60.47 & 56.11 & 39.45 & 41.95 &32.17 & 44.02 \\
        \rowcolor{gray!20}
         WeakMCN &\ding{51} & 69.26 & 70.44 & 62.94 & 52.12 & 57.28 & 43.26 &54.71 \\
         \midrule
        \multicolumn{9}{l}{Semi-RES (1\% labeled data)}\\
        \midrule
         SemiRes~\cite{yang2024sam} &\ding{55} & 50.90 & 57.54 & 44.48 & 36.49 & 42.86 &28.58 & 34.76 \\
        \rowcolor{gray!20}
         WeakMCN &\ding{51} & 59.11 & 60.44 & 56.49 & 43.00 & 49.63 & 35.90 &45.05 \\
         \bottomrule
    \end{tabular}
    }
    \label{tab:semi-sota}
    \vspace{-1em}
\end{table}

%% file: CVPR2025/figures/Visualization_cases.tex
\begin{figure*}[t!]
    \centering
    \begin{subfigure}[b]{\textwidth}
        \centering
        \includegraphics[width=0.95\textwidth]{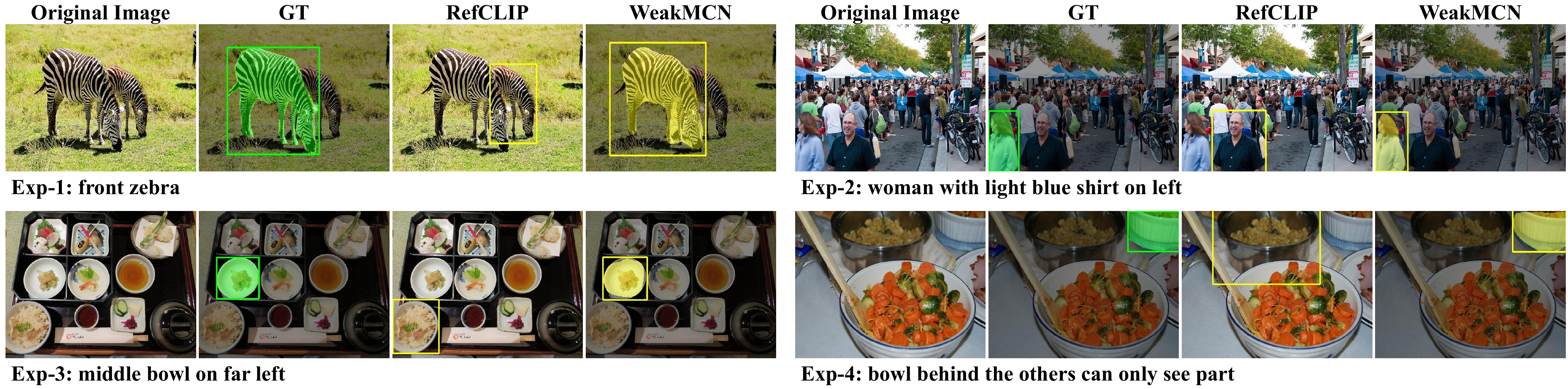}
        \caption{\textbf{Comparison of WeakMCN and RefCLIP.}}
        \vspace{-12pt}
        \label{fig:cases-sota}
    \end{subfigure}
    \vspace{-1pt}  

    \begin{subfigure}[b]{\textwidth}
        \centering
        \includegraphics[width=0.95\textwidth]{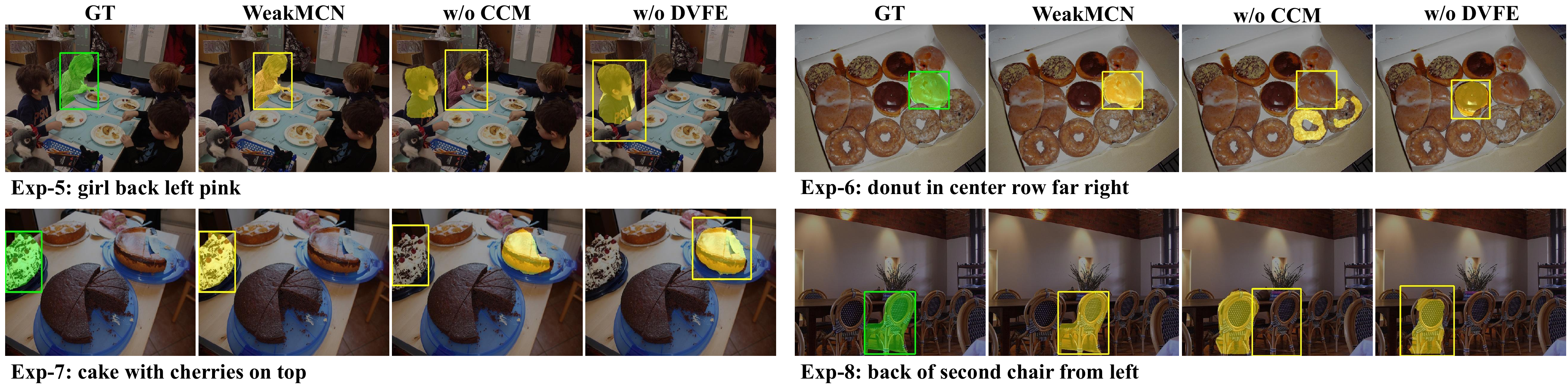}
        \caption{\textbf{Visualization of different ablation modules.}}
        \vspace{-12pt}
        \label{fig:cases-ablation}
    \end{subfigure}
    \vspace{-1em}  

    \caption{ \textbf{Visualizations of the prediction by the proposed WeakMCN.} We compare the results of WeakMCN with RefCLIP in (a) and compare the effects of our design in (b).}
    \label{fig:visualization}
    \vspace{-2em}
\end{figure*}


%% file: CVPR2025/sections/5_conclusion.tex
\section{Conclusion}

In this paper, we have proposed WeakMCN, a novel weakly supervised multi-task network for WREC and WRES. WeakMCN unifies these traditionally separate tasks under weak supervision, achieving effective multi-task learning through innovative feature enhancement and consistency mechanisms. Specifically, our DVFE module adaptively combines diverse visual features, while the CCM promotes alignment between detection and segmentation outputs. Together, these components ensure effective collaboration between WREC and WRES, resulting in significant performance gains. Extensive experiments on multiple benchmarks demonstrate WeakMCN’s superiority over existing methods in both tasks. Moreover, our approach exhibits strong generalization capabilities in both semi-supervised and weakly supervised scenarios.



%% file: CVPR2025/sections/6_acknowledgement.tex
\section*{Acknowledgement}
This work was supported by the National Natural Science Foundation of China (No. 623B2088), the China Postdoctoral Science Foundation (No. 2024M761548), King’s-China Scholarship Council PhD Scholarship programme (K-CSC), and Hubei Province Natural Science Foundation (No.2023AFB267).

%% file: CVPR2025/supple2.tex
\appendix
\clearpage
\section{Supplementary Materials}
In this supplementary material, we present additional qualitative and quantitative results of our proposed WeakMCN. Section ~\ref{sec:add_abl} includes ablation studies on (1) a comparative analysis between our trainable WRES head and a straightforward SAM-based pipeline, demonstrating the advantages of our approach, (2) the sensitivity analysis of the ISL threshold, (3) the impact of different visual features in DVFE, and (4) the parameter efficiency comparison with existing methods. Section ~\ref{sec:fail_case} analyzes typical failure cases to identify current limitations and future directions for improvement.



\subsection{Additional Ablation Studies}
\label{sec:add_abl}
\input{CVPR2025/tables/Baseline_comparison_v2}
\noindent \textbf{Comparison with Direct SAM Application.}
Our method leverages SAM for generating pseudo masks to train the segmentation head. An alternative strategy is to directly employ SAM for mask generation at inference time.  To quantitatively evaluate these two approaches, we conducted comparative experiments, with results presented in Table~\ref{tab:baseline_comparison_sam}: first training a REC model with DVFE for localization (first row), then using its predicted boxes to prompt SAM for mask generation at inference time (second row).
While this pipeline achieves competitive performance, achieving 67.36\% REC and 53.97\% RES on RefCOCO, we observe a notable performance gap compared to our proposed WeakMCN (third row), particularly in RES performance. For instance, on RefCOCO, WeakMCN outperforms this alternative approach by 1.19\% and 4.18\% in REC and RES metrics respectively. The performance gap highlights two key advantages of our approach: (1) While both methods utilize SAM, ours leverages it only for pseudo mask generation during training, allowing our lightweight WRES head to learn task-specific features, whereas direct SAM application is entirely dependent on the quality of the predicted bounding boxes of WREC head at inference time. (2) Our trainable WRES head enables dynamic feature interaction with the WREC head during training, fostering mutual enhancement between WREC and WRES. These results validate our design choice of using SAM as a teacher model for training rather than as a direct inference tool.


\noindent \textbf{The impact of the threshold in ISL.}~Tab.~\ref{tab:consis} presents the impact of varying hyperparameter thresholds $\alpha$ in ISL. For RefCOCO, the best performance is observed at $\alpha = 0.3$, achieving improvements of 0.61\% and 0.19\% in the WREC and WRES tasks, respectively, compared to the worst-performing configuration. Similarly, for RefCOCO+, the optimal performance occurs at $\alpha = 0.2$, with gains of 1.29\% and 0.91\% in the WREC and WRES tasks, respectively. Overall, these results demonstrate that the proposed WeakMCN model exhibits robustness to the choice of $\alpha$, showing minimal sensitivity to this hyperparameter. In this paper, we adopt $\alpha = 0.3$ for consistency across experiments.

\input{CVPR2025/tables/Consistency_check}
\input{CVPR2025/tables/sup_visualbank}
\input{CVPR2025/tables/para_flops_inference}
\input{CVPR2025/figures/sup_failcase}
\noindent \textbf{More visual features in visual bank.} To investigate the impact of incorporating additional visual features into our model, we conduct detailed ablation studies on the Dynamic Visual Feature Encoder (DVFE) as shown in Table~\ref{tab:rich_vision_clip}. We systematically evaluate three visual features: DINO features ($V_{dino}$), SAM features ($V_{sam}$), and CLIP features ($V_{clip}$). Our experiments reveal that while the combination of $V_{dino}$ and $V_{sam}$ achieves strong performance, further incorporating $V_{clip}$ leads to slight performance degradation. For example, on RefCOCO, we observe performance drops of 0.41\% and 0.51\% for REC and RES tasks respectively when adding $V_{clip}$ to the $V_{dino}$+$V_{sam}$ combination. We hypothesize that this degradation stems from the redundant information and training noise introduced by excessive visual features, which may contaminate the learned feature representations. This finding emphasizes the crucial importance of maintaining a balanced and efficient visual feature bank rather than merely accumulating features.

\noindent \textbf{The efficiency of DVFE.} 
As shown in Table~\ref{tab:backbone_eff}, we conduct ablation studies to analyze the efficiency-performance trade-off of our proposed DVFE. 
The baseline model with only DarkNet features ($V_{dark}$) achieves 24.5 FPS but shows limited performance (63.95\% REC, 46.88\% RES on RefCOCO). 
By incorporating DINO features ($V_{dino}$), the inference speed slightly decreases to 20.3 FPS, while bringing substantial improvements in both REC (+3.42\%) and RES (+9.26\%). 
The full DVFE implementation with all three features ($V_{dark}$, $V_{dino}$, and $V_{sam}$) further boosts the performance to 68.55\% REC (+4.60\% over baseline) and 58.15\% RES (+11.27\% over baseline) on RefCOCO, at the cost of reducing inference speed to 17.7 FPS. 
Similar performance gains are observed on RefCOCO+, where the full DVFE achieves significant improvements in both REC (+11.65\%) and RES (+12.86\%) compared to using $V_{dark}$ alone. 
These results demonstrate that while additional features moderately impact computational efficiency, the performance benefits of our DVFE are substantial and justify the modest decrease in inference speed. 
The flexible architecture of DVFE enables different feature combinations to meet various speed-accuracy requirements in real-world applications.

\noindent \textbf{Efficiency Comparison with SOTA Methods.} 
The experimental results in Table~\ref{tab:params} demonstrate the comprehensive advantages of our WeakMCN in terms of parameter efficiency, training efficiency, and inference speed. 
From the perspective of model size, with only 34.31M trainable parameters, WeakMCN significantly reduces the number of learnable parameters by 31.3\%, 76.5\%, and 69.8\% compared to APL (49.91M), Shatter (145.96M), and TRIS (113.56M), respectively. 
In terms of training efficiency, WeakMCN requires only 7 hours for convergence, which is considerably faster than Shatter (25.5h) and comparable to APL (7.5h). 
For inference speed, WeakMCN achieves 17.7 FPS, showing better real-time capability than APL (18.2 FPS) and significantly outperforming Shatter (7.51 FPS). 
Despite being more efficient, WeakMCN achieves state-of-the-art performance on both tasks, surpassing RefCLIP (60.36\%) by 8.19\% and APL (64.51\%) by 4.04\% in REC accuracy (68.55\%), while outperforming Shatter (34.76\%) by 23.39\% and TRIS (31.17\%) by 26.98\% in RES performance (58.15\%). 
Particularly noteworthy is that WeakMCN is the only model that simultaneously handles both REC and RES tasks while maintaining competitive efficiency metrics. 
These results validate the effectiveness of our multi-task learning framework in achieving a superior balance between computational efficiency and performance enhancement.


\subsection{Failure Cases}
\label{sec:fail_case}

Fig.~\ref{fig:supfailurecase} illustrates typical failure cases that reveal the current limitations of our approach. Specifically, cases 1-3 demonstrate that WeakMCN tends to produce oversegmented predictions when multiple objects overlap within a single detected bounding box, despite achieving accurate localization. Furthermore, cases 4-6 showcase the model's difficulty in processing complex and lengthy expressions, particularly in terms of precise object localization. These failure cases indicate that there remains substantial room for improvement in WeakMCN's visual reasoning capabilities and scene understanding, especially for handling intricate spatial relationships and complex visual contexts.
\label{sec:more_exp_details}


%% file: CVPR2025/tables/Baseline_comparison_v2.tex

\begin{table}[h]
    \centering
    \caption{\textbf{Comparison of replacing the WRES head with the SAM head.}}
    \resizebox{0.45\textwidth}{!}{
    \begin{tabular}{l|cc|cc}
        \toprule
         \multirow{2}{*}{Model} & \multicolumn{2}{c|}{RefCOCO} & \multicolumn{2}{c}{RefCOCO+} \\
         &REC&RES &REC &RES \\
         \midrule
         WeakMCN (w/o WRES) &67.36 &- &48.94 &-\\
         \midrule
         WeakMCN (w/o WRES) + SAM$_\text{head}$ & 67.36 & 53.97 &48.94 &37.97\\
         WeakMCN &68.55 &58.15 &51.48 &41.48\\
        \bottomrule
    \end{tabular}
    }
    \label{tab:baseline_comparison_sam}
\end{table}

%% file: CVPR2025/tables/Consistency_check.tex
\begin{table}[t]
    \centering
    \caption{\textbf{Comparison of various hyperparameter thresholds ($\alpha$) in ISL.}}
    \resizebox{0.3\textwidth}{!}{
    \begin{tabular}{c|cc|cc}
        \toprule
         \multirow{2}{*}{$\alpha$} & \multicolumn{2}{c|}{RefCOCO} & \multicolumn{2}{c}{RefCOCO+} \\
         &REC&RES &REC &RES \\
         \midrule
         0.1 &68.03 &57.82 &50.26 &41.58\\
         0.2 &68.38 &57.91 &51.48 &41.48\\
         0.3 &68.55 &58.15 &50.49 &41.34\\
         0.4 &68.64 &58.03 &50.19 &40.57\\
        \bottomrule
    \end{tabular}
    }
    \label{tab:consis}
    \vspace{-1em}
\end{table}

%% file: CVPR2025/tables/sup_visualbank.tex
\begin{table}[t]
    \centering
    \caption{\textbf{Ablation studies of DVFE in WeakMCN.}}
    \resizebox{0.45\textwidth}{!}{
    \begin{tabular}{ccc|cc|cc}
        \toprule
         \multicolumn{3}{c|}{$\mathcal{B}$} & \multicolumn{2}{c|}{RefCOCO} & \multicolumn{2}{c}{RefCOCO+} \\
         $V_{dino}$&$V_{sam}$ &$V_{clip}$&REC&RES &REC &RES \\
         \midrule
         \ding{51} & & &67.37 &56.14 &50.32 &40.43 \\
         \ding{51}&\ding{51} & &68.55 &58.15 &51.49 &41.47 \\
          \ding{51}&\ding{51} & \ding{51} & 68.14 & 57.64 & 50.98 & 40.76 \\
        \bottomrule
    \end{tabular}
    }
    \label{tab:rich_vision_clip}
\end{table}

%% file: CVPR2025/tables/para_flops_inference.tex



\begin{table}[h]
    \centering
    \caption{\textbf{The efficiency of DVFE in WeakMCN.}}
    \resizebox{0.45\textwidth}{!}{
    \begin{tabular}{ccc|c|cc|cc}
        \toprule
         \multicolumn{3}{c|}{Features in DVFE} &\multirow{2}{*}{Infrence Speed.}& \multicolumn{2}{c}{RefCOCO} & \multicolumn{2}{c}{RefCOCO+} \\
         $V_{dark}$&$V_{dino}$&$V_{sam}$ &&REC&RES &REC&RES\\
         \midrule
        \ding{51} & & & 24.5fps & 63.95 &46.88 & 39.84 & 28.61 \\
        \ding{51} & \ding{51} & & 20.3fps &67.37 &56.14 & 50.32 & 40.43 \\
        \ding{51} & \ding{51} & \ding{51} & 17.7fps &68.55 &58.15 & 51.49 & 41.47 \\
        \bottomrule
    \end{tabular}
    }
    \label{tab:backbone_eff}
\end{table}

\begin{table}[t]
    \centering
    \caption{\textbf{Comparison of parameters with other weakly-supervised RES or REC methods.} Params denote the number of trainable parameters. Train denotes training hours. Inf denotes inference speed.}
    \resizebox{0.45\textwidth}{!}{
    \begin{tabular}{l|c|c|c|c|cc|cc}
        \toprule
         \multirow{2}{*}{Model} & \multirow{2}{*}{Multi-task} &
         \multirow{2}{*}{Params (M)} &
         \multirow{2}{*}{Train (h)} &
         \multirow{2}{*}{Inf (fps)} &
         \multicolumn{2}{c|}{RefCOCO} & \multicolumn{2}{c}{RefCOCO+} \\
         & & & & &REC&RES &REC &RES \\
         \midrule
         RefCLIP~\cite{jin2023refclip} & \ding{55} & 27.50 & 5 & 31.3 &60.36 &- &40.39 &-\\
         APL~\cite{luo2025apl} & \ding{55} & 49.91 & 7.5 & 18.2 & 64.51 & - & 42.70 & -\\
         TRIS~\cite{liu2023referring} & \ding{55} & 113.56 & - & - & - &31.17 &- &30.90\\
         Shatter~\cite{kim2023shatter} & \ding{55} & 145.96  & 25.5 & 7.51& - &34.76 &- &28.48\\
         \rowcolor{gray!20}
         WeakMCN & \checkmark & 34.31 & 7 &17.7 &68.55 &58.15 &51.48 &41.48\\
        \bottomrule
    \end{tabular}
    }
    \label{tab:params}
    \vspace{-1em}
\end{table}

%% file: CVPR2025/figures/sup_failcase.tex
\begin{figure*}[t!] \centering
\includegraphics[width=\linewidth]{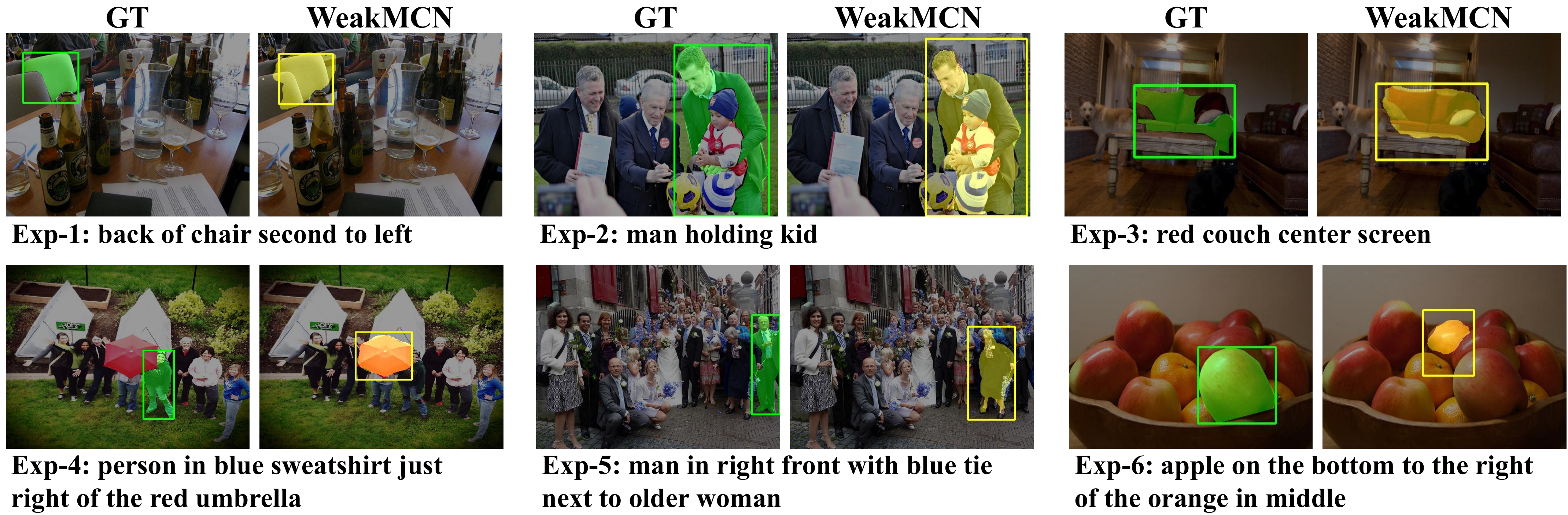}
\caption{\textbf{Failure cases.} The green mask/bounding box is the ground truth, and the yellow one is our prediction.}
\label{fig:supfailurecase}
\end{figure*}